\documentclass[letterpaper, 10 pt, conference]{ieeeconf}
\IEEEoverridecommandlockouts
\usepackage{cite}
\usepackage{booktabs}
\usepackage{url}

\usepackage[pdftex]{graphicx}
\usepackage{subfig}
\graphicspath{{./figures/}}
\usepackage{amsmath}
\usepackage{nth}
\usepackage{cases}

\usepackage{tikz}
\usetikzlibrary{shapes, arrows}

\usepackage{varwidth}
\usepackage{caption}

\title{\LARGE \bf Extrinsic Calibration of 3D Range Finder and Camera without Auxiliary Object or Human Intervention
}

\author{Qinghai Liao$^{1}$, Ming Liu$^{1}$, Lei Tai$^{2}$, Haoyang Ye$^{2}$
\thanks{$^{*}$This work was supported by the Research Grant Council of Hong Kong SAR Government, China, under project No. 16206014 and No. 16212815; National Natural Science Foundation of China No. 6140021318, awarded to Prof. Ming Liu}
\thanks{$^{1}$Qinghai Liao and Ming Liu are with Department of Electronics and Computer Engineering Engineering,
Hong Kong University of Science and Technology.
        {\tt\small qinghai.liao@connect.ust.hk, eelium@ust.hk}}        
\thanks{$^{2}$ Lei Tai and Haoyang Ye are with Department of Mechanical and Biomedical Engineering, City University of Hong Kong.
        {\tt\small lei.tai@my.cityu.edu.hk, hy.ye@my.cityu.edu.hk}}
}

\begin{document}
\maketitle
\thispagestyle{empty}
\pagestyle{empty}
\begin{abstract}
Fusion of heterogeneous extroceptive sensors is the most effient and effective way to representing the environment precisely, as it overcomes various defects of each homogeneous sensor. The rigid transformation (aka. extrinsic parameters) of heterogeneous sensory systems should be available before precisely fusing the multisensor information. Researchers have proposed several approaches to estimating the extrinsic parameters. These approaches require either auxiliary objects, like chessboards, or extra help from human to select correspondences. In this paper, we proposed a novel extrinsic calibration approach for the extrinsic calibration of range and image sensors. As far as we know, it is the first automatic approach with no requirement of auxiliary objects or any human interventions.

First, we estimate the initial extrinsic parameters from the individual motion of the range finder and the camera. Then we extract lines in the image and point-cloud pairs, to refine the line feature associations by the initial extrinsic parameters.

At the end, we discussed the degenerate case which may lead to the algorithm failure and validate our approach by simulation. The results indicate high-precision extrinsic calibration results against the ground-truth.
\end{abstract}

%\begin{IEEEkeywords}
%visual tracking, parameter calibration, plane object
%\end{IEEEkeywords}

% \begin{center} 
% \bfseries EDICS Category: 3-BBND 
% \end{center}
% \fi
%\IEEEpeerreviewmaketitle

% For example, the vision sensor camera can't work in the dark scene and lack scale information which drives us to combine it with a range finder.
\section{Introduction}

%\subsection{Motivation}

\subsection{Heterogeneous extroceptive sensory. Why and What?}
\label{sec:heter}

In recent years, 3D reconstruction and mapping system have got increasing attention since these techniques have wide potential applications, such as in robotic (SLAM, navigation), autonomous driving vehicle, gaming (VR) and so on. These tasks have demands on the environment model which contains rich information and strong describing ability.

However, it is either imprecise or requires high computation power to model the environment with homogeneous sensing modalities like cameras. Cameras are so far the most widely used sensors which offer a wealth of information on color, but it is heavily affected by the light and is not able to work in many scenes. People are aware of that vision sensor's flaw has indeed caused bit problems such as some autonomous driving accidents \cite{tesla}. An important solution is to combine various sensing modalities to enhance the model. The common practice is to use the 3D range finder and camera together to make complementarity. To efficiently use the information from heterogeneous modalities, we have to fuse the information and present them in a single reference frame. The goal is to compute the rigid transformation including the relative rotation and translation of different sensor coordinate systems. The 6-DoF rigid transformation is called \textit{extrinsic parameters} and the process to estimate the extrinsic parameters is called \textit{extrinsic calibration}.

The extrinsic calibration of range-image sensors was introduced by Zhang \cite{Zhang2004} with a 2D range finder and a camera. As the cost decreasing and the development of technique, 3D range finder has become popular and more extrinsic calibration research has been focusing on the 3D range finder and camera recently.

%\subsubsection{Automatic Calibration without Auxiliary Objects or Human Intervention}
\subsection{Challenges}
\label{sec:ve}

We indeed believe most fundamental theoretical problems have been solved for the extrinsic calibration, along with the development of theories in multi-view geometry \cite{hartley2003multiple} and optimization \cite{kummerle2011g}, However, the integration of algorithms and automated calibration in practical cases persists problematic and still highly challenging.
In these cases, the demands to facilities or human interventions should be minimized. By this work, we would like to tackle several last-mile problems to enable automatic \textit{feature association} and \textit{robust regression}, leading to high-precision extrinsic calibration over heterogeneous sensors.

Feature association is to find the correspondences from image pixels captured by cameras to the point-cloud captured by range finders. If we have the correspondences, this problem will be further solved as a PnP problem \cite{lepetit2009epnp}, optimization problem \cite{kummerle2011g} or even active calibration \cite{xie2015online} which have been already well-studied. However, images and point-cloud are hard to be matched due to the inherent representation difference: The images captured from cameras are \textit{dense} representations, for which each pixel has a proper definition. Point clouds are \textit{sparse} representation. Regardless the density, the space between any two observed points has no definition at all. Thereby, there are no such generic feature descriptions across heterogeneous sensors that we can directly use to match the features from images with that from point-clouds. Researchers have proposed many methods to find the correspondence and derived extrinsic calibration approaches. We categorize these methods into two categories:

\textbf{With auxiliary objects.} Firstly, extrinsic calibration techniques can rely on the external artificial calibration objects which should be observed simultaneously from the camera and camera. More than one views from different poses are necessary to perform the extrinsic calibration. The most-used calibration object is the checkerboard which is also widely used for camera intrinsic calibration \cite{Zhang2000}. Zhang\cite{Zhang2004} first published the extrinsic calibration approach which used the checkerboard to calibrate the 2D range finder and camera. Barreto \textit{et al.} \cite{Barreto2012} presented another approach by freely moving the checkerboard. They convert the problem to registering a set of planes and lines in the 3D space and get the solution by solving a standard p3p problem with a linear system. Unnikrishnan \textit{et al.}\cite{Unnikrishnan2005} extended Zhang's approach to 3D range finder with a checkerboard and their approach is the first method for a 3D range finder and camera. Pandey \textit{et al.}\cite{Pandey2010} further extended above approach to omnidirectional Camera. Rodriguez \textit{et al.}\cite{RodriguezF.2008} presented an extrinsic and intrinsic calibration approach by circle based calibration target. Beside,  Choi\cite{Choi2016} used two orthogonal planes; Aliakbarpour\cite{Aliakbarpour} used a pattern and IMU sensor together to jointly reduce the needed number of points. Gong\textit{et al.}\cite{s130201902} used an arbitrary trihedron to assist the calibration. Since an auxiliary object is required, it is hard to perform on-line recalibration for a working system.

\textbf{With human interventions.} The second category is to get the feature correspondences by manual selection but provide primary filtering to narrow down the candidates. These methods don't require artificial calibration target any more but used the features from the natural scene. Scaramuzza \textit{et al.}\cite{Scaramuzza2007} first try this way by converting the visually ambiguous 3D range information into a 2D map where natural features of a scene are highlighted. After manually selecting the correspondences the problem is a standard PnP problem. Moghadam \textit{et al.} \cite{Moghadam2013} selected all 3D lines and 2D lines as candidates for human selection which makes the algorithm more robust and precise since it largely reduces the chances that human makes a wrong decision. These methods require massive human attention for feature associate and not suitable for an on-line system as well.

The methods in the first category are easy to find constraints and get a precise solution. Nevertheless, the weakness is also obvious that placing the calibration object will modify the scene and limit the application scope. The second category makes the problem much easier to find the constraints. But with the human intervention, the algorithms lose the automation ability and the precision drops very fast due to imprecise guidance from humans. Furthermore, these existing approaches share a common drawback that they cannot work on-the-fly for on-line systems. Note that it is easy to cause variant extrinsic parameters during the test course in a real environment, such as vibration and non-rigid sensors system (typically for the autonomous driving system). It is not easy for these methods to adapt.

Pandey \textit{et al.} \cite{Pandey2012} proposed a calibration object-free extrinsic calibration method for a 3D range finder and camera by using the mutual information framework. Their approach owns the ability to do calibration during working process. However, they require the laser range finder reflectivity values which are not always reliable in practical scenes according to our tests \cite{pomerleau2012noise}.

\subsection{Contributions}
In this paper, we propose a novel automated extrinsic calibration algorithm for a 3D range finder and a camera without auxiliary objects. Our proposed algorithm separates the calibration process into two steps: initial coarse extrinsic parameter calculation and extrinsic parameter refinement. Our approach requires the sensor system to move around (rotation is required) to several poses to acquire information which can be used to figure out the initial coarse extrinsic parameter. The second step we exploit the lines in both the 2D and 3D space and find correspondences across the two representations aided by the initial parameters. These line correspondences are used to refine the extrinsic parameter. Besides, an open-ware library including the code and test scenarios has been released with the publication of the paper at \texttt{http://ram-lab.com/download}. 

 We address the following contributions of this work: 
 \begin{itemize}
 \item We present a novel extrinsic calibration algorithm for camera images and 3D point-clouds without requirements of auxiliary objects or human intervention. 
\item A robust weighted least square solution is proposed for the problem, which can be used as a generic solution for similar problems with sensory outliers.
 \item Principles of filtration and degenerated cases were studied for the proposed framework as hints for the application of the algorithms.
 \end{itemize}

\subsection{Organization}
\label{sec:orgnize}
The rest of paper is organized as follows. Section {\uppercase\expandafter{\romannumeral2}} presents the methodology of our proposed algorithm including above mentioned two steps. Section {\uppercase\expandafter{\romannumeral3}} shows the calibration results. Finally, we make conclusion and discussion in Section {\uppercase\expandafter{\romannumeral4}}.

\section{methodology}
The proposed algorithm includes initial extrinsic parameter calculation and extrinsic parameter refinement. In the first step, we make the sensor system random move (rotation required) to several poses to acquire information, based on which we calculate a set of coarse extrinsic parameters. Then we exploit the line feature in two observation space to get good matches which are the constraint to refine the extrinsic parameters. We will discuss these two steps in the following subsection.

\subsection{Initial parameter calculation}
\begin{figure}[!ht]
\centering
\includegraphics[width=3.5in]{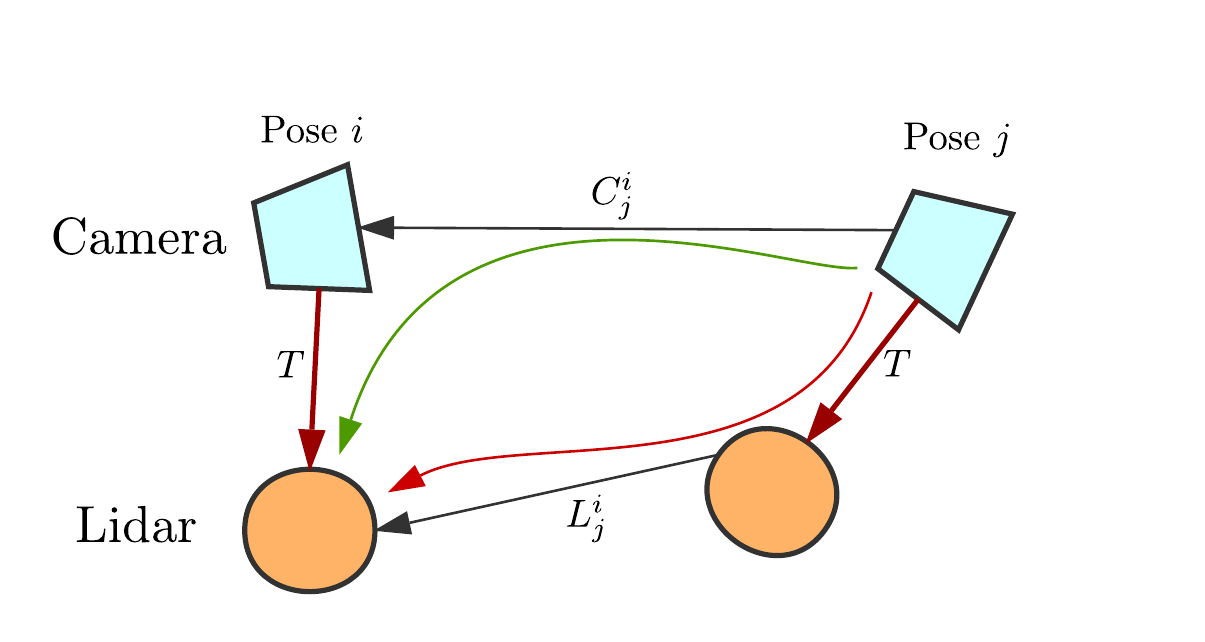}
\caption{Transformation between lidar and camera. $L_j^i$ is the relative transform computed from ICP; $C_j^i$ is the transform optimized by local bundle adjustment.}
\label{fig:rotation chain}
\end{figure}

Cameras are bearing sensors. Considering laser range finder can be also regarded as a superior bearing sensor, which is enhanced by depth measurements, we adopt the similar approach for multi-camera rig calibration without overlapping, as described in \cite{Esquivel2007, heng2013camodocal}. We assume that the collected pairs of data are taken with $N$ poses. We define the extrinsic parameters as transformation $T$:
\begin{equation}
T=
\begin{bmatrix}
R &t\\
0 &1
\end{bmatrix}
\end{equation}
$R$ is the rotation 3x3 matrix and $t$ is the translation 3x1 vector. We denote the local reference transformation from reference frame to target frame as $\mathbf{S}^{target}_{reference}$, where $\mathbf{S}:=\{L,C\}$ is the sensor type. e.g., $L^0_1$ represents the transformation from lidar's pose \#1 to pose \#0 and $C^0_1$ for that of camera. We can get at most $\mathbf{C}^N_2$ pairs \footnote{$\mathbf{C}$ is the combination operator.} as shown in Fig. \ref{fig:rotation chain} from $N$ poses. For a perspective camera, lots of pairs will be invalid since in these pairs camera does not share much overlapped field of view, which makes it difficult to get reliable relative transformation. Every pair has two poses as shown in Fig. \ref{fig:rotation chain}.

\subsubsection{Relative-pose estimation}
To calculate the extrinsic parameters, we first estimate the transformation of the lidar and the camera for each paired data, respectively. 

For point-cloud data we use our previously proposed library \texttt{libpointmatcher} \cite{Pomerleau12comp}, which includes a fast and reliable ICP implementation with point-to-plane error formulations. The ICP output indicates the rotation and translation between to consequent point-cloud scans. For image pairs, we use ORB\cite{1302} feature and g2o \cite{5979949} to perform a local two-view bundle-adjustment optimization to get the transformation up to scale. Thereby, we can compute a full transformation matrix of the lidar by point-clouds, but for camera we obtain
\begin{equation*}
T=
\begin{bmatrix}
R & \lambda t\\
0 &1
\end{bmatrix}
\end{equation*}
where $\lambda$ is the scale factor.

\subsubsection{Rotation and Translation}
In every pair the transformation between a camera pose $j$ and lidar with pose $i$ can be computed as routes in two alternative paths along the quadrilateral edges. Either by first transforming to lidar at the same pose (red curve), where
\begin{equation}
T^1 = L^i_jT
\end{equation}
or alternatively by first changing to previous camera reference frame (green curve), where
\begin{equation}
T^2 = TC^i_j
\end{equation}

$T^1$ and $T^2$ are ideally the same, i.e. 
by Eq. 2 and Eq. 3, the geometrical constraint is:
\begin{equation}
L^i_jT = TC^i_j
\end{equation}
and it must hold for every pair ($\mathbf{C}^2_N$). In Eq. 4, only the extrinsic parameter $T$ is unknown and $L^i_j$ and $C^i_j$ have been computed. Eq. 4 can be decomposed into two parts. The first part contains only rotation:
\begin{equation}
{_rL}^i_jR = R\,({_rC}^i)_j
\end{equation}
The second part contains both rotation and translation:
\begin{equation}
{_rL}^i_jt + {_tL}^i_j =\lambda^i_j R\,({_tC}^i)_j+t
\end{equation}

Here the scale factor $\lambda$ of camera is unknown yet. $_rL$ is the rotation matrix within $L$ and $_tL$ denotes the translation in $L$;  $_rC$ is the rotation matrix within $C$ and $_tC$ denotes the translation in $C$.

The common practice to solve Eq. 4 is to solve the Eq. 5 first and solve Eq. 6 with the result $R$ from Eq. 5. For Eq. 5, Chou \textit{at el.} \cite{Chou1991} have presented the rotation with quaternion which reduce the variable number from 9 to 4. The Eq. 5 can be re-written as:
\begin{equation}
q_lq=qq_c
\end{equation}
or equivalently 
\begin{equation}
(T_{q_l}-T^*_{q_c})q=0
\end{equation}
where $T_q$and $T^*_q$ are defined as follows. By letting  $q=(w, x, y, z)^T$,
\begin{equation*}
T_{q}=
\begin{bmatrix}
w & -x & -y &-z \\
x & w & -z& y \\
y & z & w& -x \\
z & -y & x& w
\end{bmatrix}
\end{equation*}
\begin{equation*}
T^*_{q}=
\begin{bmatrix}
w & -x & -y &-z \\
x & w & z& -y \\
y & -z & w& x \\
z & y & -x& w
\end{bmatrix}
\end{equation*}

Hence we will get the linear equation system with $N$ pairs for the rotation $q$ subject to $|q|=1$.
\begin{equation}
A
\begin{bmatrix}
w  \\
x \\
y \\
z 
\end{bmatrix}
=
\begin{bmatrix}
0  \\
0 \\
0 \\
0 
\end{bmatrix}
\end{equation}
where $A$ is (4N)x4 coefficient matrix
\begin{equation*}
\footnotesize
\arraycolsep=3pt
A=
\begin{bmatrix}
w^0_l - w^0_c & -x^0_l + x^0_c  & -y^0_l + y^0_c  &-z^0_l + z^0_c \\
x^0_l - x^0_c & w^0_l - w^0_c   & z^0_l - z^0_c   & -y^0_l + y^0_c \\
y^0_l - y^0_c & -z^0_l +z^0_c   & w^0_l - w^0_c   & -x^0_l - x^0_c\\
z^0_l - z^0_c & y^0_l - y^0_c   & -x^0_l + x^0_c  & w^0_l -w^0_c\\
\vdots & \vdots & \vdots & \vdots \\
w^{4N}_l - w^{4N}_c & -x^{4N}_l + x^{4N}_c  & -y^{4N}_l + y^{4N}_c  &-z^{4N}_l + z^{4N}_c \\
x^{4N}_l - x^{4N}_c & w^{4N}_l - w^{4N}_c   & z^{4N}_l - z^{4N}_c   & -y^{4N}_l + y^{4N}_c \\
y^{4N}_l - y^{4N}_c & -z^{4N}_l +z^{4N}_c   & w^{4N}_l - w^{4N}_c   & -x^{4N}_l - x^{4N}_c\\
z^{4N}_l - z^{4N}_c & y^{4N}_l - y^{4N}_c   & -x^{4N}_l + x^{4N}_c  & w^{4N}_l -w^{4N}_c
\end{bmatrix}
\end{equation*}
Eq. 9 subjects to $|q|=1$ and SVD is an efficient tool to get the $q$.

After recovery the rotation, Eq. 6 could be re-formulated as 
\begin{equation}
\begin{bmatrix}
I-_rL^i_j & R\,(_tC^i_j)
\end{bmatrix}
\begin{bmatrix}
t\\
\lambda^i_j
\end{bmatrix}
=_tL^i_j
\end{equation}
The fully expansion of above equation will be 
\begin{equation}
\footnotesize
\arraycolsep=3pt
\begin{bmatrix}
I-_rL^0 & R(_tC^0)&0&\hdots&0\\
I-_rL^1 & 0 &R(_tC^1)&\hdots&0\\
\vdots & \vdots & \vdots & \vdots &\vdots \\
I-_rL^N & 0 &0 &\hdots&R(_tC^N)
\end{bmatrix}
\begin{bmatrix}
t\\
\lambda^0\\
\vdots\\
\lambda^N
\end{bmatrix}
=
\begin{bmatrix}
_tL^0\\
_tL^1\\
\vdots\\
_tL^N
\end{bmatrix}
\end{equation}
Here we replace the index $i,j$ with pair index to make it clear. Eq. 11 has 3+N variable and 3N constraints which means it can be solved with at least two pairs (three poses). 

\subsubsection{Filtration of big error pairs}
\begin{figure}[!ht]
\centering
\includegraphics[width=3.5in]{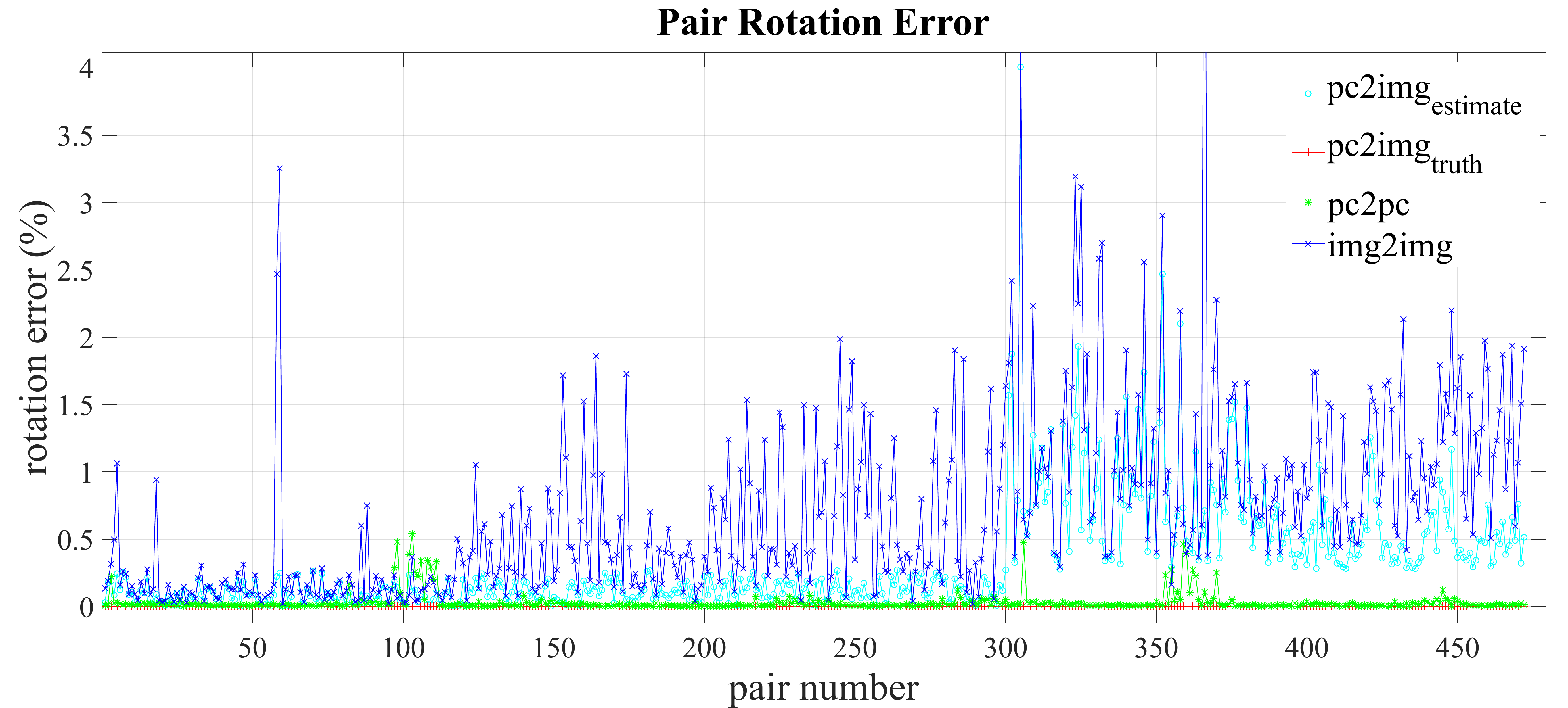}
\caption{Rotation error of camera(blue) and lidar(green), quaternion rotation angle between camera and lidar of estimation(cyan) and ground truth(red) }
\label{fig:pair_rotation_error}
\end{figure}

If we have accurate transformation of camera and lidar above proposed algorithm is enough to get the extrinsic parameter of this system. However, the accurate transformation is hard to recover, especially for camera. Due to the limited filed of view (FOV), intrinsic parameter error, feature extraction and mismatching and so on, compared with $360^{\circ}$ FOV lidar the camera transformation estimation usually has low quality. As shown in Fig. \ref{fig:pair_rotation_error}, camera rotation estimation has maximum 7\% error\footnote{roattion error is defined as $err=\frac{acos(|q_1{\cdot}q_2|)}{pi/2}$} and lidar's error is less than 1\%. 

Inaccurate and unreliable camera transformation makes the initial parameter fragile. To improve the accuracy of initial parameter we have to filter out some pairs which introduce big errors that usually caused by camera transformation. Hence, we need the principle of filtration for this purpose. As we represent Eq. 5 in quaternion format 
\begin{equation*}
\label{eq:1}
\begin{aligned}
p^lq &= qp^c  \\
w^l &= w^c
\end{aligned}
\end{equation*}
we can get $\theta^l=\theta^c$ where $\theta$ is the quaternion rotation angle which means that no matter what the extrinsic parameter is the ration angles of lidar and camera are always identity in every motion. And as shown in Fig. \ref{fig:pair_rotation_error} the ground truth of rotation angle between lidar and camera is zeros but estimated data instead. The rotation angle between lidar and camera from estimated transformation(cyan) nearly follows the camera transformation error line(blue) which proves the above assertion. Hence, we can take the rotation angle as principle of filtration and set the threshold.

\subsubsection{Degenerate cases}
Above proposed algorithm has two degenerate cases. The first one is easy to realize that the sensor system's motion must include rotation. When the sensor system does purely translation motion Eq. 5 will degenerate and has no constraint on rotation. 

The second case occurs when the sensor system rotates against the same one axis. In this case, Eq. 7 is an identical equation and the solution of $q$ is arbitrary. 

Both cases will cause the rotation failure and  consequently, make Eq. 6 unsolvable and break off the algorithm. However, it is not hard to avoid these two cases in practices.

\subsection{parameter refinement}
The previous step gets the initial extrinsic parameter which has high quality estimation of rotation. But the translation of camera estimated from image often has poor accuracy and the scale factor helps the error-propagation. The final result is the unacceptable translation error and this is the purpose of this step.

In this paper we propose to use the natural scene line feature to do refinement. $Line^{3D}-line^{2D}$ correspondence could derive strong constraint on direction and line-to-line distance which correspondingly refine the rotation and translation, respectively. %Since initially estimated rotation is accurate and to balance the complexity and speed we simplify this step to translation refinement with vertical lines. 

\subsubsection{2D line extraction}
\begin{figure}[!ht]
\centering
%\includegraphics[width=2.6in]{image_lines}
%\includegraphics[width=2.6in]{image_lines_filtering}
%\caption{2D lines detection and filtering, above show all detected lines, below show the filtering effect}
\includegraphics[width=3.5in]{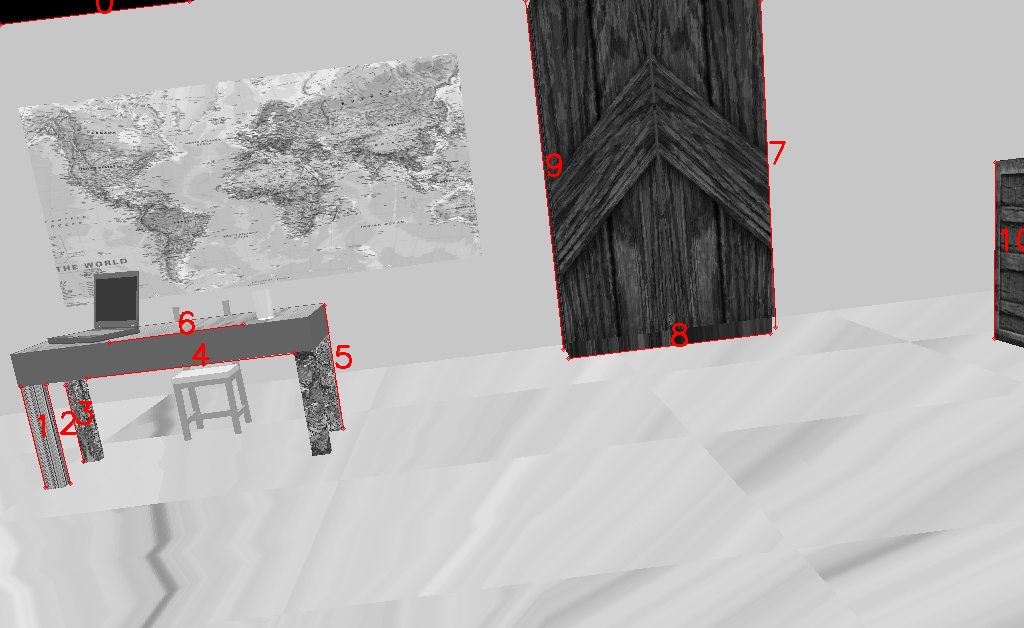}
\caption{2D lines detection}
\label{fig:2d line}
\end{figure}

We utilize Akinlar \textit{at el.}\cite{Akinlar:2011:ERL:2027478.2027695} proposed EDlines lines detection approach for image lines extraction. According to the configuration of line length, anchor, gradient threshold, fit error threshold we can easily find the all lines candidate. Fig.\ref{fig:2d line} shows the result in one pose.

\subsubsection{3D line extraction}
3D arbitrarily lines detection in point cloud is difficult and slow especially for unorganized point cloud. To make the proposed algorithm work on-the-fly we focus more on the 3D lines detection speed. We accept certain false detected result since initial extrinsic estimation could help remove them during matching process.

\begin{figure}[!ht]
\centering
\includegraphics[width=3.4in]{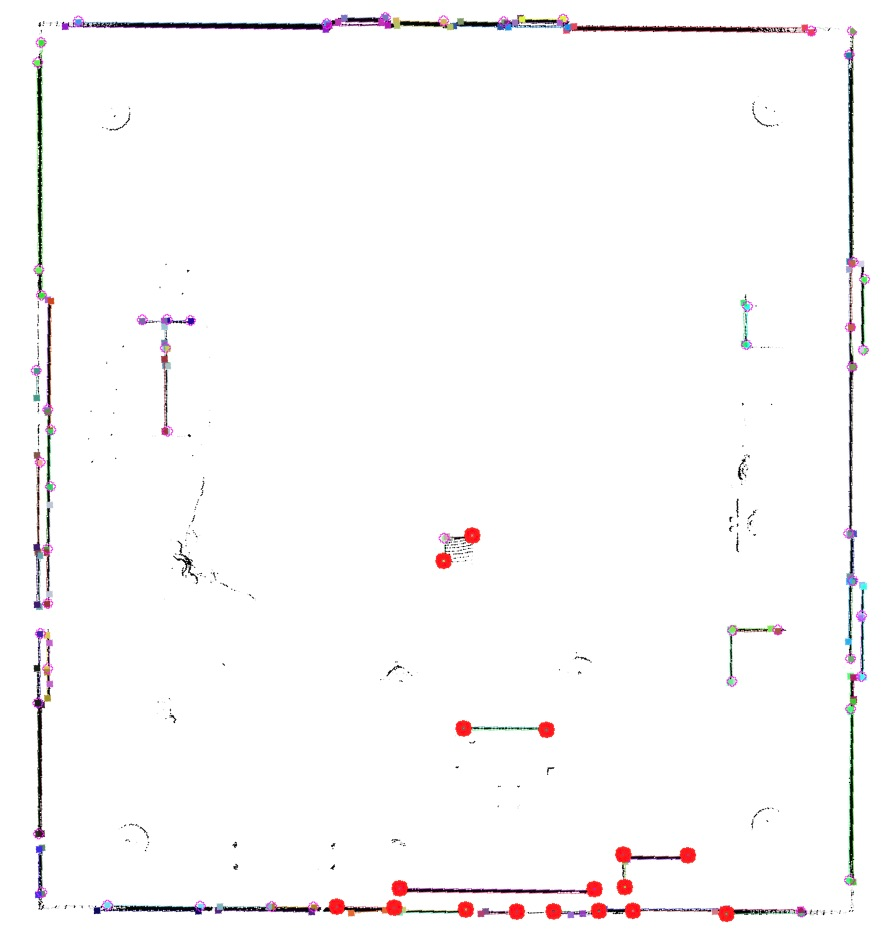}
\caption{3D lines detection,end points(color dots), filtration result(big red dot)}
\label{fig:3d line}
\end{figure}

Our proposed algorithm uses simple vertical lines whose direction vector is $[0 \quad 1 \quad 0]^T$(vertical axis is Y-axis in our experiment). We propose a speed and simple vertical line detection that projects all point to floor surface and the number of points is regard as the image intensity. Later we apply line detection method on the projection intensity image to find all end-points which we think has great chance to be the vertical line. As shown in Fig.\ref{fig:3d line} second figure, we highlight the detected lines with different color and their end-points.

\subsubsection{Line matching}
3D line and 2D line matching are based on the initial extrinsic parameter estimation. We find the correspondences by two threshold of angle error and point-to-line distance. Since we use the vertical lines in lidar space hence all desired lines should have constant direction. Let vector $dz$ be the third column of $R^{-1}$ and project it into image coordinate by camera intrinsic parameter K. The new 2D line's direction should be the truth value. We filter the 2D lines according to this value and the example is shown in Fig.\ref{fig:2d line} second figure.
For 3D lines, we mainly remove points that are not in the FoV of camera and the remaining (big red dot in Fig.\ref{fig:3d line}) end-point are the candidate 3D line features.

We present 2D line as $w_0u+w_1v+w_2=0$. End-point $[x \quad 1 \quad z]^T$ corresponding to 3D line can be prjected to image as 
\begin{equation}
z\begin{bmatrix}
u'  \\
v' \\
1  
\end{bmatrix}
=K(R^{-1}
\begin{bmatrix}
x  \\
1 \\
z
\end{bmatrix}-R^{-1}t)
\end{equation}
The error can be define as format of normal point-to-line distance without normalization
\begin{equation}
err = \begin{bmatrix}
w_0 & w_1 &w_2
\end{bmatrix}
\begin{bmatrix}
u'  \\
v' \\
1  
\end{bmatrix}
\end{equation}
When $err$ is smaller than the threshold we classify them as valid correspondences.

\subsubsection{Robust Least Square Solution with Huber Weights}
Since outliers and noise are ubiquitous for the observation of 2D line, 3D line, and corners. Proper rectification of the result is necessary. We aim to minimize 
a global penalty function, which is a sum of the re-projection error. A generic form of the penalty function is:
\begin{eqnarray*}
&\text{minimize}  &\phi(r_1) + \phi(r_2) + \cdots + \phi(r_m)\\
&\text{subject to} & \vec{r} = Ax-b
\end{eqnarray*}
where $A\in \mathbf{R}^{mxn}$ is the configuration matrix, which in this case the collections of matching constraints, $\phi: \mathbf{R} \rightarrow \mathbf{R}$ is a convex penalty function.
$r$ is the remainder. 
Note that $\phi(\cdot)$ is often taking a quadratic form, i.e. $\phi(u) = u^2$, namely Ordinary Linear Square (OLS) as used by most g2o implementations. However, such a quadratic form will introduce high sensitivity
to outliers. Regularization can partially solve the over-fitting caused by the outliers, but the strength of the error introduced by the outliers persist \cite{kolter2009regularization}. The problem can also be partially solved by quartile regression \cite{koenker2005quantile} but with high computational and sampling cost. In this work, we use a weighted robust
least square (RLS) with Huber weights, as:
\begin{equation}
  \label{eq:huber}
  \phi_{hub}(u) = 
\left\{ 
\begin{aligned}
1\cdot u^2, &\, \text{for} |u|\le M\\
\frac{|u|}{M} \cdot u^2, &\, \text{for} |u|>M
\end{aligned}
\right.
\end{equation}
It results in a new convex penalty function $\sum_I \phi_{hub}^i(u_i)$, where $i$ is the index of the observation. $M$ is empirically selected as 3. Note that this weighting function 
highlights the supporting data which can fit the computed $x$, while the effect of the outliers drops from quadratic to linear form. It leads to enhanced performance in
pose estimation. 
%A statistic of the error in x, y, z directions and in terms of error vector norm is shown as table \ref{my-label}.
%\begin{table}[h]
%\centering
%\caption{A comparative result among estimations}
%\label{my-label}
%\begin{tabular}{llll}
%\toprule
%     & Initial Guess & OLS        & RLS(proposed)        \\
%\midrule
%dx   & 0.014251      & 0.0020824  & 0.0098112  \\
%dy   & 0.157605      & 0.0506278  & -0.0010439 \\
%dz   & -0.014592     & -0.0079447 & -0.0059059 \\
%norm & 0.15892       & 0.051290   & 0.011499  \\
%\bottomrule
%\end{tabular}
%\end{table}
\section{Experiment}
We validate the proposed algorithm in V-rep simulation environment \cite{vrep} which offers the out-of-box sensors and models. In addition, simulation environment gives out precise ground truth which is hard to measure in real environment. However, it also has drawbacks like the vision sensor gets information of the scene with much less texture than real environment.

\begin{figure}[!ht]
\centering
\includegraphics[width=2.6in]{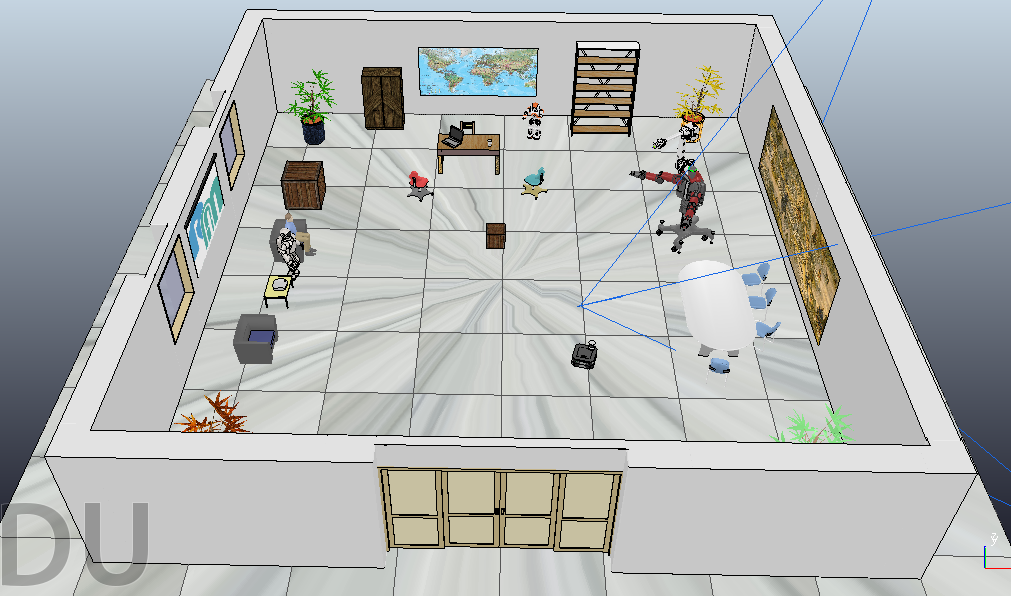}
\caption{V-rep simulation scene}
\label{fig:vrep scene}
\end{figure} 

\begin{figure}[!ht]
\centering
\includegraphics[width=3in]{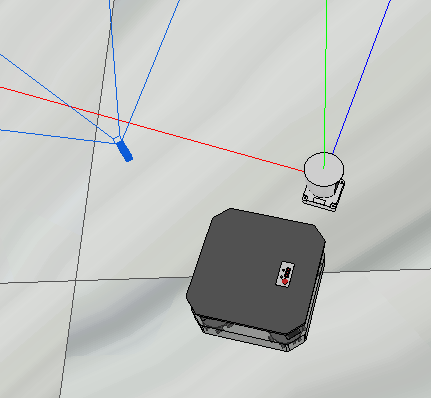}
\caption{sensor configuration, blue is vision sensor, grey is 3D range finder}
\label{fig:vrep sensor}
\end{figure}

Fig.\ref{fig:vrep scene} shows our experiment simulation scene. It's a quite complex scene but has great resemblance with mobile robot working place. Sensor configuration is shown in Fig.\ref{fig:vrep sensor}. We attach the vision sensor and 3D range finder to the omnidirectional car with fixed extrinsic parameter. The car randomly moves (including rotation and translation) in the room and sensors acquire the information simultaneously. During this process we keep the system's motion away from the degenerate cases.

\begin{figure}[!ht]
\centering
\subfloat[Rotation error]{\includegraphics[width=3.4in]{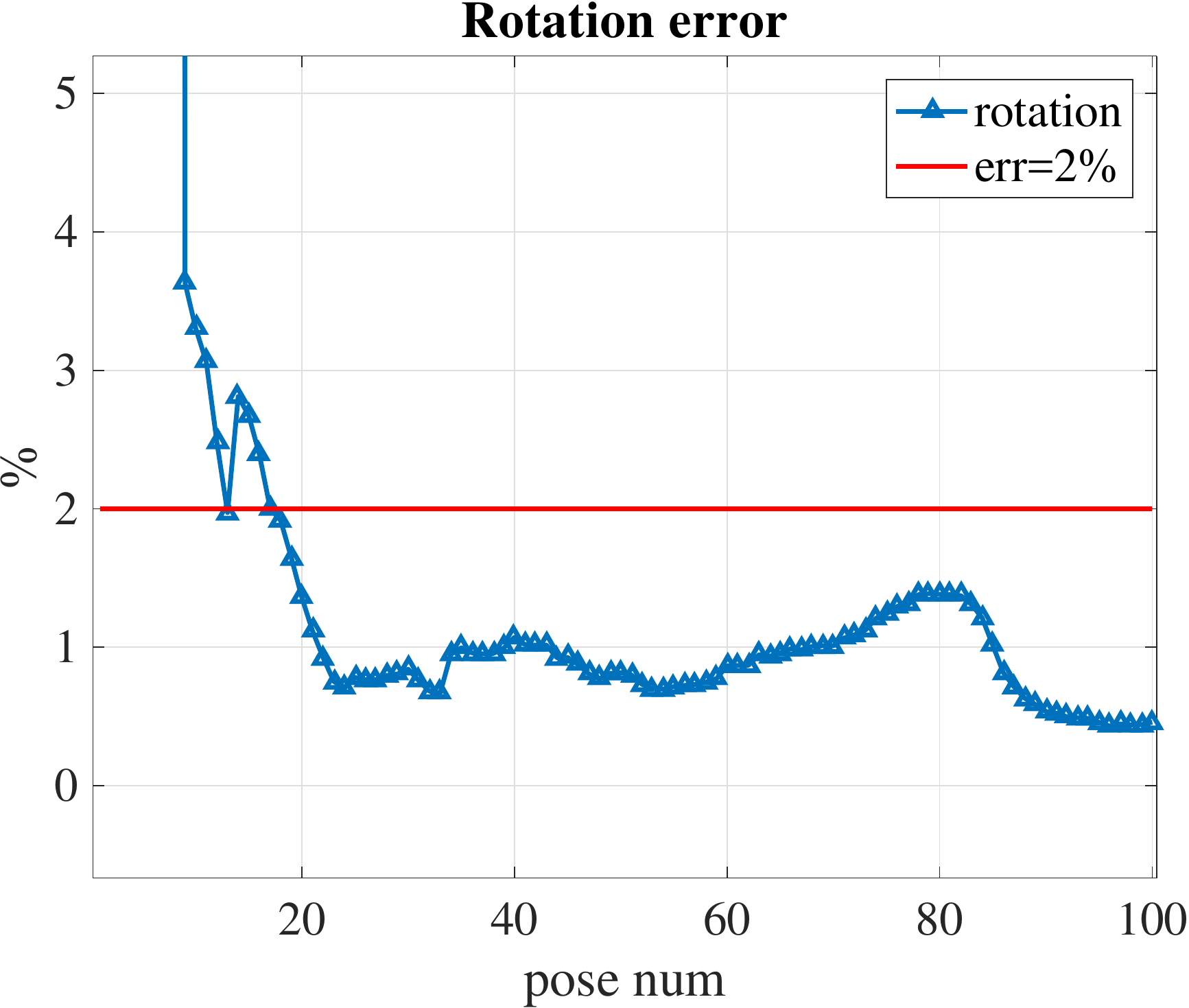}
\label{fig:error_initial_rotation}}
\hfil
\subfloat[Translation error, y axis is the lidar's vertical direction]{\includegraphics[width=3.4in]{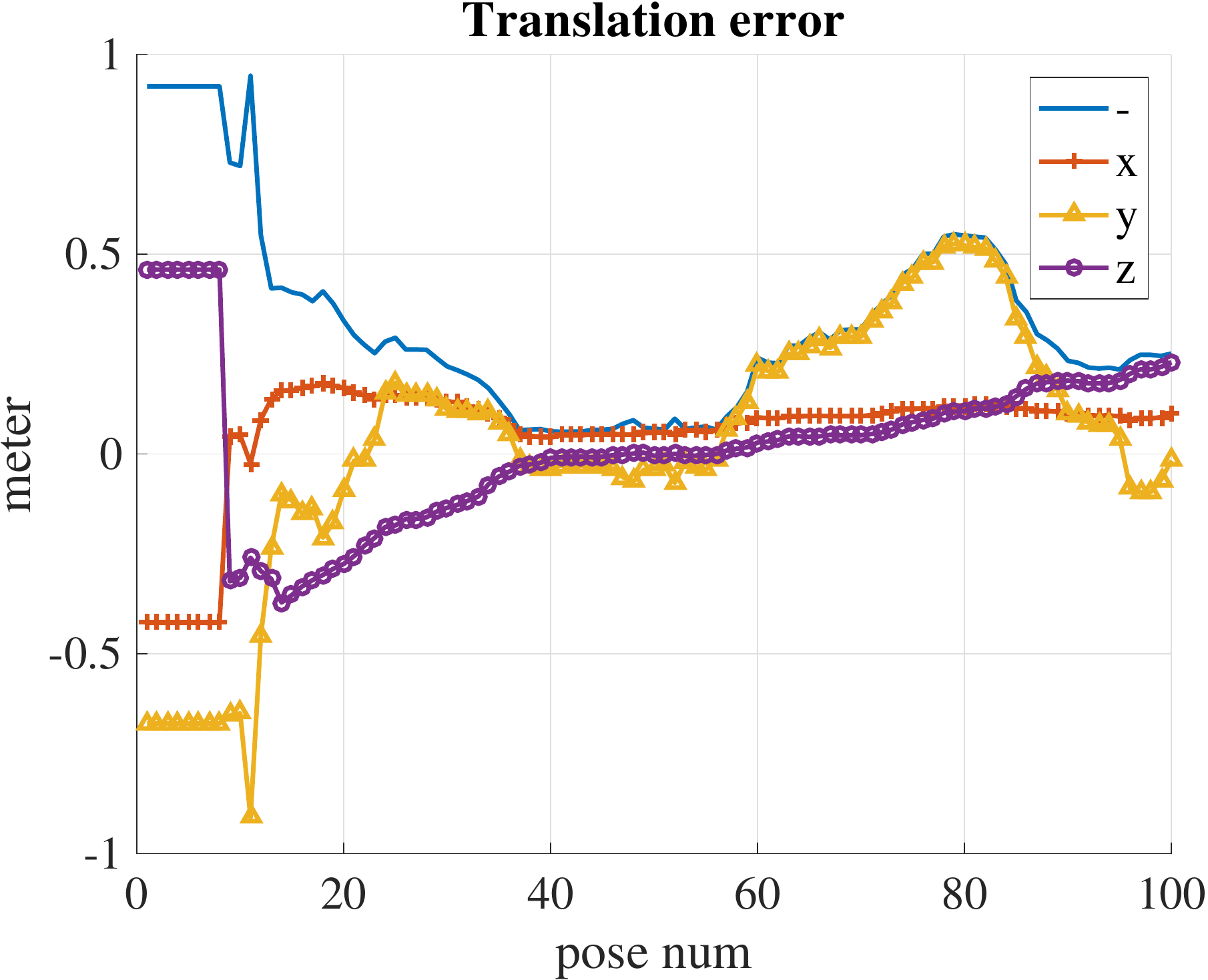}
\label{fig:error_initial_translation}}
\caption{Error changes with pose number after filtering invalid pairs}
\label{fig:error_initial}
\end{figure}

\begin{figure}[!ht]
\centering
\includegraphics[width=3.2in]{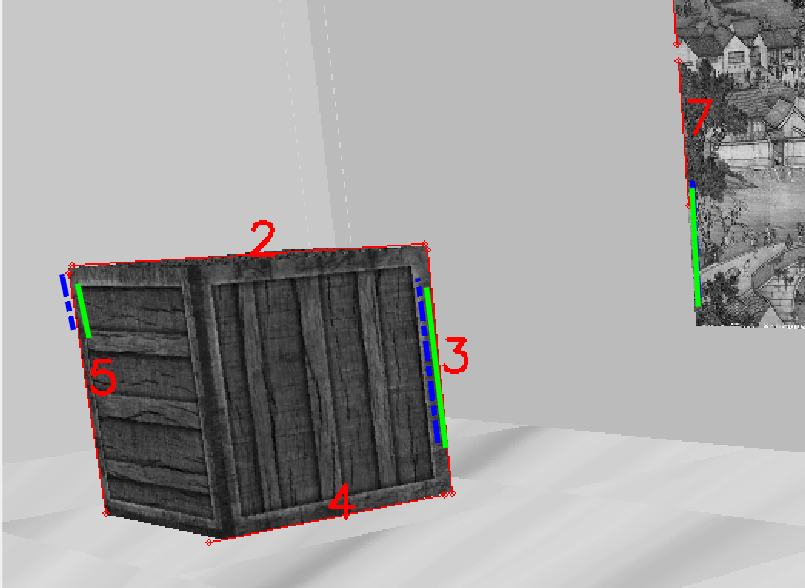}
\caption{Line re-projection, image 2D lines (red), projected 3D lines with extrinsic parameter of ground truth (green) and initial process (blue). }
\label{fig:line_match}
\end{figure}

As shown in Fig. \ref{fig:error_initial} and Fig. \ref{fig:line_match}, our initial calculation algorithm works very well, especially for rotation estimation. In Fig. \ref{fig:error_initial_rotation}, big rotation error exists when pose number is lower than 10 since we set the threshold for the pose number for launching the algorithm. For the translation, we may note the error increment in y-axis as $pose number \in [60,80]$. In this paper, we take the lidar frame as world frame and the y-axis is the lidar's vertical direction which relatively has lower resolution than horizontal plane. From Fig. \ref{fig:line_match} we can see that the initial calculation algorithm result is accurate enough for the line matching algorithm which is able to lead fewer mismatching.

\begin{table}[!ht]
\centering
\caption{Experiment results in rotation}
\label{tab:result rotation}
\renewcommand\arraystretch{1.1}
\renewcommand\tabcolsep{2pt}
\begin{tabular}{lllll}
\toprule
     & Rotation & Error(deg) \\
\midrule
Ground Truth   & [-0.0394   -0.4220    0.0764   -0.9025]      & -  \\
Initial(no filtering)   & 
[-0.8955    0.0438    0.4399    0.0514]     & 1.5680  \\
Initial(filtering)   & [-0.0394   -0.4220    0.0764   -0.9025]     & 0.0034  \\
\bottomrule
\end{tabular}
\end{table}

\begin{table}[!ht]
\centering
\caption{Experiment results in translation}
\label{tab:result translation}
\renewcommand\arraystretch{1.1}
\renewcommand\tabcolsep{2pt}
\begin{tabular}{lllll}
\toprule
     & Translation & Error (m) & Ratio\\
\midrule
Ground Truth   & [0.4224    0.6745   -0.4616]     & - & 0\%  \\
Initial(no filtering)   & [ 0.3992    0.1861   -0.3964
]    & 0.4932  & 53.60\%\\
Initial(filtering)   & [0.4082    0.5168   -0.4470
] & 0.1589  & 17.27\%\\
Refinement(OLS)&[0.4203    0.6238   -0.4536]&0.05129&5.57\%\\
Refinement(proposed)& [0.4322    0.6734   -0.4675]&0.0115&1.25\%\\
\bottomrule
\end{tabular}
\end{table}

Table \ref{tab:result rotation} and Table \ref{tab:result translation} shows the experiment result. Rotation and translation are both validated. Especially, the filtering and refinement process largely improve the accuracy of the proposed algorithm.

%Ming: why no comparison to existing methods.. need justify
Note that the error of the resulted calibrated translation and rotation is comparatively lower than state-of-the-art approaches, even for cases with auxiliary objects \cite{xie2015online}, even with highly drifted initial conditions. Due to limited space, he full comparison results will be presented in a separate report in \url{http://ram-lab.com/file/report_extrinsic.pdf}. 

\section{Conclusion}
In this paper we presented an extrinsic calibration approach for heterogeneous extroceptive sensors. It is independent of auxiliary objects or human interventions, which largely relief the requirement comparing with existing approaches. We also provided a validated calibration framework, which is online available at \url{http://ram-lab.com/download}. Despite the limitation that the method should work in environments with detectable lines, it still by-far the most convenient plug-and-play extrinsic calibration system to the community.

For future work, we will extract more robust 3D lines and refine the rotation and translation simultaneously with close-form solutions. We are also looking into direct association methods between camera images and point-clouds.


%\section*{Acknowledgment}
%This work is supported by the Research Grant Council of Hong Kong SAR
%Government, China, under project No. 16206014 and No. 16212815; National
%Natural Science Foundation of China No. 6140021318, awarded to Prof.
%Ming Liu.



%\bibliographystyle{IEEEtran}
%\begin{thebibliography}{1}
%Franois CHaumette and Seth Hutchinson, ``Visual Servo Control Part I: Basic Approaches," IEEE Robot. Automat. Mag. vol. 13, no. 4, pp. 82–90, 2006.
%  
%\end{thebibliography}

\bibliographystyle{IEEEtran}
%\bibliography{IEEEabrv,exCalibration}
\bibliography{exCalibration}
\end{document}